\documentclass[letterpaper, 10 pt, conference]{ieeeconf}  

\IEEEoverridecommandlockouts                              

\usepackage[hidelinks]{hyperref} 
\usepackage{graphicx}
\usepackage{cite}
\usepackage{float}
\usepackage{amsmath,amssymb,amsfonts}
\usepackage[table]{xcolor}
\usepackage{threeparttable}
\usepackage{xcolor}
\definecolor{rose}{RGB}{199,21,133} 
\usepackage{booktabs}
\usepackage{tabularx}
\usepackage{multirow}
\usepackage{array}
\usepackage{amssymb}
\newcommand{\cmark}{\ensuremath{\checkmark}}

\title{\LARGE \bf
Grounded in Time: A Multi-Source Dataset and Benchmark for Temporal Grounding in Robotic Manipulation
}

\author{\
Yi Wang$^{1\dagger}$, Yang Yang$^{1\dagger}$, Guangqi Xu$^{1}$, Sumin Lin$^{1}$, Ning Kang$^{1}$, \
Pengxiang Lu$^{1}$, Xiaotong Chen$^{1}$,\\\\\
Zeyu Xue$^{1}$, Ping Deng$^{3}$, Xing Liu$^{4}$, Chenguang Yang$^{2\*}$ and Zhenyu Lu$^{1\*}$
\thanks{\textsuperscript{1}H.I.-Robot Lab., School of Automation Science and Engineering, South China University of Technology.}
\thanks{\textsuperscript{2}Department of Computing, The Hong Kong Polytechnic University, Hong Kong, 999077, HKSAR.}\
\thanks{\textsuperscript{3}Faculty of Mechanical and Aerospace Engineering, The Hong Kong University of Science and Technology, HKSAR.}\
\thanks{\textsuperscript{4}School of Astronautics, Northwestern Polytechnical University, Xi'an, China.}\
\thanks{\textsuperscript{$\dagger$}Yi Wang and Yang Yang contributed equally to this work.}\
\thanks{\*Corresponding author}
}

\makeatletter

\long\def\IEEEaftertitletext#1{%
    \def\@IEEEaftertitletext{#1}%
}

\newcommand{\teasercaption}[1]{%
    \refstepcounter{figure}%
    \@makecaption{\fnum@figure}{#1}%
}

\makeatother

\begin{document}

\IEEEaftertitletext{%
\begin{minipage}{\textwidth}

    {\centering
    \includegraphics[width=\linewidth]{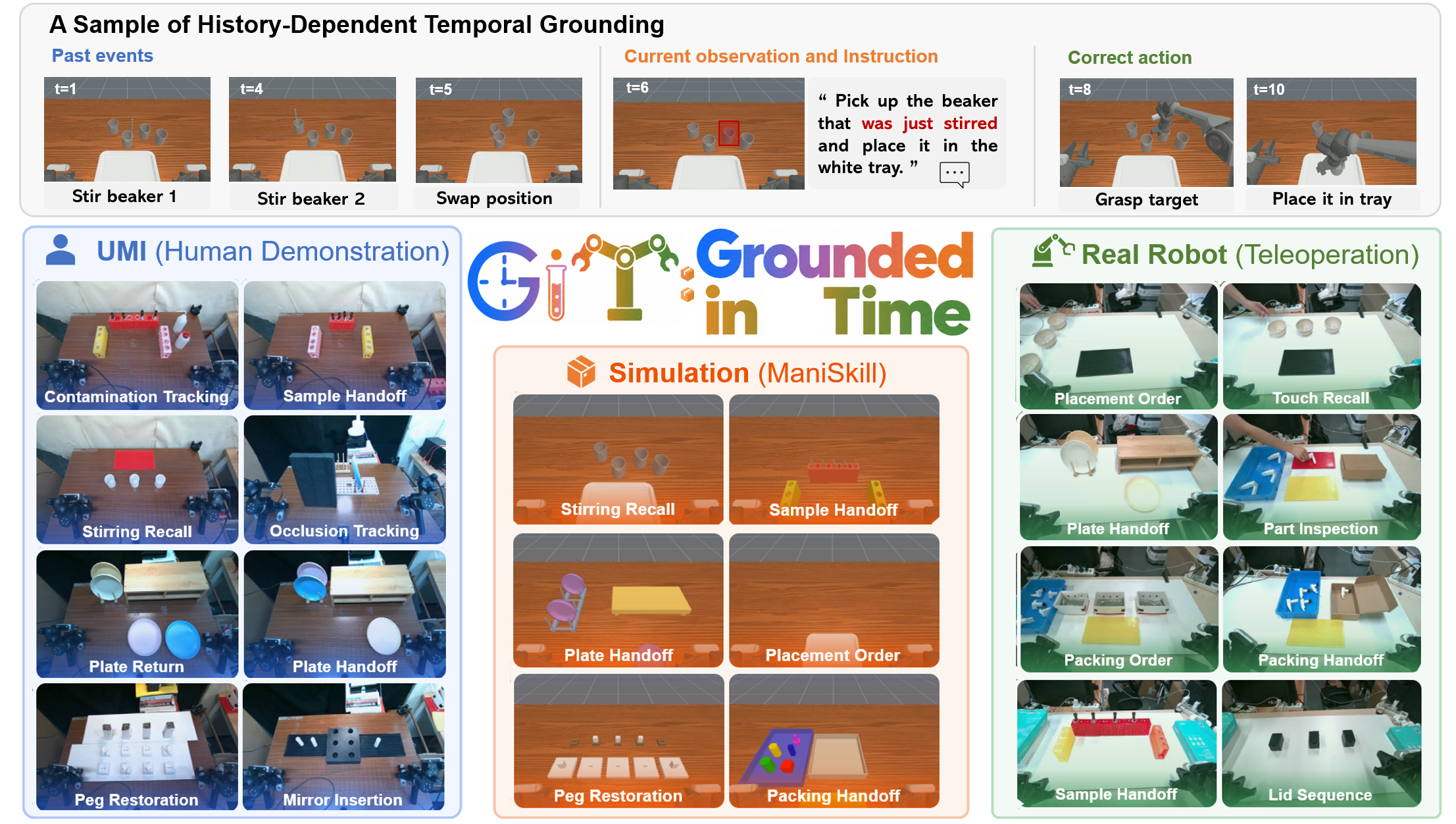}\par
    }

    \teasercaption{Overview of GiT, a multi-source dataset and benchmark for episodic event memory in robotic manipulation. GiT comprises 18 history-conditioned manipulation tasks spanning BioLab, Household, and Industrial scenarios, with data acquired through UMI-based human demonstration, ManiSkill simulation, and VR-based real-robot teleoperation. Representative task instances cover interaction-state recall, persistent object-identity tracking, temporal-order reasoning, cross-arm handoffs, and history-guided state restoration.}
    \label{fig:git-overview}

\end{minipage}%
}

\thispagestyle{empty}
\maketitle
\pagestyle{empty}

\begin{abstract}
Robotic manipulation often requires inferring task-relevant states from past interactions when the current observation alone is insufficient to determine the appropriate action. Despite progress in benchmarking memory-augmented vision-language-action (VLA) models, application-oriented tasks requiring history-dependent semantic inference remain underrepresented. We introduce GiT (Grounded in Time), a dataset and benchmark for grounding manipulation decisions in past events across biolaboratory, household, and industrial scenarios. It includes real-robot and Universal Manipulation Interface (UMI) style demonstrations covering 18 bimanual tasks, together with simulation data and a ManiSkill-based evaluation suite covering nine tasks. Fine-grained subtask annotations and annotated counterfactual task pairs, in which similar current observations require different actions depending on prior events, support policy learning and targeted evaluation of history use. Evaluations of representative end-to-end VLA models in simulation and on selected real-world tasks reveal substantial room for improvement in history-dependent manipulation. The dataset and benchmark are available at \href{https://kk-stephen.github.io/grounded-in-time/}{\textcolor{rose}{the project page}}.
\end{abstract}

\newcolumntype{C}{>{\centering\arraybackslash}X}

\begin{table*}[t]
\centering
\caption{Compact comparison of representative robotic manipulation benchmarks.}
\label{tab:benchmark}
\renewcommand{\arraystretch}{1.08}
\setlength{\tabcolsep}{3pt}
\footnotesize
\setlength{\aboverulesep}{0pt}
\setlength{\belowrulesep}{0pt}
\setlength{\extrarowheight}{0.65ex} 
\begin{tabular}{
    l
    c
    >{\raggedright\arraybackslash}p{2.6cm}
    c c c c c c c
}
\toprule
\textbf{Dataset} &
\textbf{Year} &
\textbf{Focus} &
\textbf{Env.} &
\textbf{Embodiment} &
\textbf{Scale} &
\textbf{Hist.} &
\textbf{Pre-act.} &
\textbf{Snap.} &
\textbf{CF} \\
\midrule
BEHAVIOR-1K\cite{li2023behavior} & 2024 & Daily tasks & Sim / real & Mobile arm & 1k tasks & $\times$ & $\times$ & $\checkmark$ & $\times$ \\
RoboEval\cite{wang2025roboeval} & 2025 & Bimanual tasks & Sim & Dual arm & 8 families; 3k demos & $\times$ & $\times$ & $\checkmark$ & $\times$ \\
RoboMME\cite{dai2026robomme} & 2025 & Memory tasks & Sim / real & Single arm & 16 tasks; 1.6k demos & in-episode/video & $\checkmark$ & $\times$ & $\times$ \\
RMBench\cite{chen2026rmbench} & 2026 & Memory tasks & Sim / real & Dual arm & 9 tasks; 3 real tasks & in-episode & $\checkmark$ & $\times$ & $\times$ \\
\midrule
\rowcolor{gray!20}
GiT benchmark (ours) & 2026 & Memory tasks & Sim / real & Dual arm & 18 tasks; 4.6k demos & in-episode/video & $\checkmark$ & $\times$ & $\checkmark$ \\
\bottomrule
\end{tabular}
\centering

\vspace{6pt}
\parbox{0.96\textwidth}{
\scriptsize\strut
\textit{Abbreviations:} 
\textbf{Hist.}: history dependence;
\textbf{Pre-act.}: pre-action context;
\textbf{Snap.}: snapshot sufficiency;
\textbf{CF}: counterfactual pairs.
}
\end{table*}

\section{INTRODUCTION}

Recent advances in vision-language models (VLMs) have strengthened visual understanding and instruction following \cite{liu2024improved}. Building on these advances, vision-language-action (VLA) models have demonstrated generalizable manipulation across tasks and environments\cite{shao2025large}, including long-horizon household activities \cite{kim2025openvla,black2025pi05}. In such long-horizon settings, current observations may be insufficient for action planning, motivating memory-augmented policies that support spatial recall, action sequencing, and task-progress tracking \cite{fang2025sam2act,shi2026memoryvla}. Within this broader effort to leverage interaction history, we focus on a specific capability crucial for following history-dependent instructions: associating past interaction events with objects and task-relevant states in the current scene to determine appropriate actions. For example, when several bowls look alike, executing ``Put the bowl I just used back in the cupboard'' requires identifying the target from interaction history rather than appearance alone. This capability is particularly relevant to household assistance and collaborative assembly, where actions by humans or other robots can determine the referents and requirements of subsequent instructions.

Existing robotic memory-centric benchmarks, such as \cite{dai2026robomme, chen2026rmbench} focus on in-episode state tracking, primarily assessing memory through abstract, purpose-built manipulation tasks. While useful for isolating specific memory skills, these settings offer limited coverage of everyday workflows in which prior interactions determine the meaning of subsequent instructions. In such workflows, a new request may refer to interactions completed before it was issued, including those performed by humans or other robots. When the current observation provides insufficient evidence to identify the intended objects or task-relevant states, the robot must interpret these earlier events and link them to the current scene. We term this capability \textit{pre-episode semantic grounding}: grounding a current instruction in scene objects and task-relevant states using interactions completed before the current task episode begins.

To support \textit{pre-episode semantic grounding}, we introduce \textbf{Grounded in Time (GiT)}, a multi-source dataset and benchmark for history-dependent bimanual manipulation in biolaboratory, household, and industrial settings. The GiT dataset combines simulated trajectories from ManiSkill3 \cite{DBLP:journals/corr/abs-2410-00425}, real-robot demonstrations collected through virtual reality (VR) teleoperation, and human demonstrations collected with a sensorized handheld gripper inspired by the Universal Manipulation Interface (UMI) \cite{chi2024universal}. It comprises over 4,680 episodes across 18 tasks, totaling more than 3.3 million steps and more than 30 hours of interaction. Each sample pairs a pre-episode interaction video with a manipulation instruction and execution trajectory.

Both the training and test splits include annotated counterfactual A--B pairs with matched scenes at the start of the current task but different pre-episode histories and required actions. The GiT benchmark uses the test-set pairs to assess whether policies ground current instructions in the relevant prior interactions.

Our main contributions are as follows:
\begin{enumerate}
    \item \textbf{A multi-source dataset.}
    We introduce the GiT dataset, linking pre-episode interaction histories to subsequent bimanual manipulation across three application domains.

    \item \textbf{A semi-automated annotation pipeline.}
    We develop a pipeline for hierarchical trajectory and segment level annotation, covering 180 subtask categories.

    \item \textbf{A benchmark and empirical evaluations.}
    We establish the GiT benchmark and evaluate selected end-to-end VLA policies, revealing limitations in history-dependent manipulation. We complement these evaluations with small-scale real-robot experiments.
\end{enumerate}

\section{Related Work}


\subsection{Datasets for Robotic Learning}
Robotic manipulation datasets have expanded the coverage of tasks, environments, and embodiments available for policy learning. Simulation-based resources such as RLBench \cite{james2020rlbench} and LIBERO \cite{liu2023libero} provide diverse tasks with generated or human-collected demonstrations. In the real world, BridgeData V2 \cite{walke2023bridgedata} and DROID \cite{khazatsky2024droid} broaden skill and scene coverage, while Open X-Embodiment \cite{embodiment2024openx} and RoboMIND \cite{wu2025robomind} support learning across robotic embodiments. Human demonstrations provide another source of supervision: RH20T \cite{fang2024rh20t} pairs human videos with robot executions, and UMI \cite{chi2024universal} enables manipulation data collection using handheld grippers. Domain-specific efforts such as AutoBio \cite{lan2025autobio} and LabUtopia \cite{li2025labutopia} target biology-laboratory automation with high-fidelity simulation of fluid handling, instrument interfaces, and chemically meaningful interactions, emphasizing precise protocol execution rather than memory-dependent instruction grounding. These resources primarily support learning how to execute specified tasks across diverse conditions. However, greater trajectory diversity and access to demonstration videos do not by themselves ensure supervision for interpreting requests whose referents depend on earlier interactions. The GiT dataset targets the challenges by linking pre-episode interaction videos to subsequent instructions and execution trajectories, with counterfactual A--B pair annotations in both training and test splits.

\subsection{Memory‑centric benchmarks for robotic manipulation} 
Robotic manipulation benchmarks evaluate different aspects of task execution. Table \ref{tab:benchmark} summarizes a comparison of representative robotic manipulation benchmarks along key dimensions. BEHAVIOR-1K \cite{li2023behavior} emphasizes realistic, long-horizon everyday activities, while RoboEval \cite{wang2025roboeval} assesses bimanual execution quality through fine-grained behavioral metrics. However, task realism and execution quality alone do not isolate the need for historical information. To evaluate history dependence more directly, RMBench \cite{chen2026rmbench} organizes manipulation tasks by memory complexity, whereas RoboMME \cite{dai2026robomme} evaluates counting, object permanence, reference resolution, and motion imitation, including tasks that identify manipulation targets from prior videos. These controlled tasks help isolate specific memory skills, but success on such tests alone does not establish reliable grounding of history-dependent instructions in practical workflows. Compared with prior benchmarks, our benchmark uniquely targets the pre‑episode memory capability. It requires policies to reason over completed historical episodes for decision‑making, rather than maintaining and updating continuous task states within ongoing interactions. Moreover, our benchmark complements previous efforts by focusing on \textit{pre-episode semantic grounding} in application-oriented biolaboratory, household, and industrial tasks.                                                                                          
\section{Methodology}


\subsection{Problem Formulation}
We investigate \emph{temporal grounding under perceptual ambiguity} in VLA models, examining whether they can use historical demonstrations to guide subsequent manipulation decisions when the current observation and language instruction alone are insufficient to identify the instruction-relevant manipulation target. An instruction may refer to object properties, relations, or event order established through interactions completed before the current task episode. When the current scene provides insufficient evidence of these distinctions, the intended target or task requirements cannot be reliably determined from the current observation and instruction alone. For example, visually similar objects may undergo different interactions without lasting changes to their appearance and subsequently return to similar spatial configurations, such that only their distinct interaction histories reveal which object satisfies the instruction.

Formally, we formulate temporal grounding at decision step $t$ of each benchmark instance $X_i$ as a mapping from the historical visual sequence $\mathcal{H}_i$, the current observation $\mathbf{o}_{i,t}$, and the language instruction $\ell_i$ to the instruction-relevant manipulation specification $Y_i^*$:
\begin{equation}
Y_i^* = G\left(\mathcal{H}_i, \mathbf{o}_{i,t}, \ell_i\right).
\label{eq:temporal-grounding}
\end{equation}
Here, $Y_i^*$ specifies one or more target objects together with, where applicable, a required execution order, relational configuration, or target state. The function $G$ abstractly represents how this specification is determined from historical evidence, the current observation, and the language instruction, without imposing any particular model architecture.

\begin{table*}[t]
\centering
\caption{Overview of the 18 GiT tasks, acquisition-pipeline coverage,
and required history-dependent capabilities.}
\label{tab:task-overview}

\renewcommand{\arraystretch}{1.08}
\setlength{\tabcolsep}{3pt}
\footnotesize

\begin{tabular}{
    l
    c
    >{\raggedright\arraybackslash}p{5.3cm}
    c c c
    c c c c c
}
\toprule

\multirow{2}{*}{\textbf{Scenario}} &
\multirow{2}{*}{\textbf{ID}} &
\multirow{2}{*}{\textbf{Task}} &
\multicolumn{3}{c}{\textbf{Acquisition Pipeline}} &
\multicolumn{5}{c}{\textbf{History-Dependent Capability}} \\

\cmidrule(lr){4-6}
\cmidrule(lr){7-11}

& & &
\textbf{Real} &
\textbf{UMI} &
\textbf{Sim} &
\textbf{IHR} &
\textbf{OIT} &
\textbf{TOR} &
\textbf{HG} &
\textbf{HSR} \\

\midrule

\multirow{6}{*}{BioLab}
& L1 & Tube Selection by Contamination History
& \cmark & \cmark & --
& \cmark & -- & \cmark & -- & -- \\

& L2 & \textbf{Container Selection by Filling History}
& \cmark & \cmark & \cmark
& \cmark & \cmark & -- & -- & -- \\

& L3 & Placement by Wiping History
& \cmark & \cmark & --
& \cmark & \cmark & \cmark & -- & -- \\

& L4 & \textbf{Beaker Selection by Stirring History}
& \cmark & \cmark & \cmark
& \cmark & \cmark & \cmark & -- & -- \\

& L5 & \textbf{Sample Selection by Handoff History}
& \cmark & \cmark & \cmark
& \cmark & \cmark & \cmark & \cmark & -- \\

& L6 & Lid Closure by Opening Order
& \cmark & \cmark & --
& -- & -- & \cmark & -- & -- \\

\midrule

\multirow{6}{*}{Household}
& H1 & \textbf{Plate Selection by Occlusion History}
& \cmark & \cmark & \cmark
& \cmark & \cmark & -- & -- & -- \\

& H2 & Tableware Retrieval by Event History
& \cmark & \cmark & --
& \cmark & \cmark & \cmark & -- & \cmark \\

& H3 & \textbf{Bowl Selection by Placement Order}
& \cmark & \cmark & \cmark
& -- & \cmark & \cmark & -- & -- \\

& H4 & Bowl Selection by Touch History
& \cmark & \cmark & --
& \cmark & -- & \cmark & -- & -- \\

& H5 & \textbf{Plate Placement by Handoff History}
& \cmark & \cmark & \cmark
& \cmark & \cmark & -- & \cmark & \cmark \\

& H6 & Button Activation by Press Order
& \cmark & \cmark & --
& -- & -- & \cmark & -- & -- \\

\midrule

\multirow{6}{*}{Industrial}
& I1 & Part Selection by Inspection History
& \cmark & \cmark & --
& \cmark & \cmark & \cmark & -- & -- \\

& I2 & \textbf{Bin Selection by Opening History}
& \cmark & \cmark & \cmark
& \cmark & -- & \cmark & -- & -- \\

& I3 & \textbf{Peg Insertion by Removal History}
& \cmark & \cmark & \cmark
& \cmark & -- & -- & -- & \cmark \\

& I4 & Part Retrieval by Packing Order
& \cmark & \cmark & --
& \cmark & -- & \cmark & -- & -- \\

& I5 & \textbf{Part Packing by Handoff Order}
& \cmark & \cmark & \cmark
& -- & -- & \cmark & \cmark & -- \\

& I6 & Peg Insertion by Mirror Position History
& \cmark & \cmark & --
& \cmark & -- & \cmark & -- & -- \\

\bottomrule
\end{tabular}\par

\vspace{6pt}
\parbox{0.96\textwidth}{
\scriptsize
\textit{Abbreviations:} IHR, interaction-history recall;
OIT, object-identity tracking;
TOR, temporal-order reasoning;
HG, handoff grounding;
HSR, history-guided state restoration.
A check mark in an acquisition-pipeline column indicates that the corresponding
pipeline provides data for the task, whereas a check mark in a capability
column indicates that the task requires that capability. A task may require
multiple capabilities. Task names in \textbf{bold} indicate tasks covered by
both the simulation pipeline and the GiT benchmark.
}

\end{table*}

\begin{figure}[ht!]
\centering
\includegraphics [width=\linewidth]{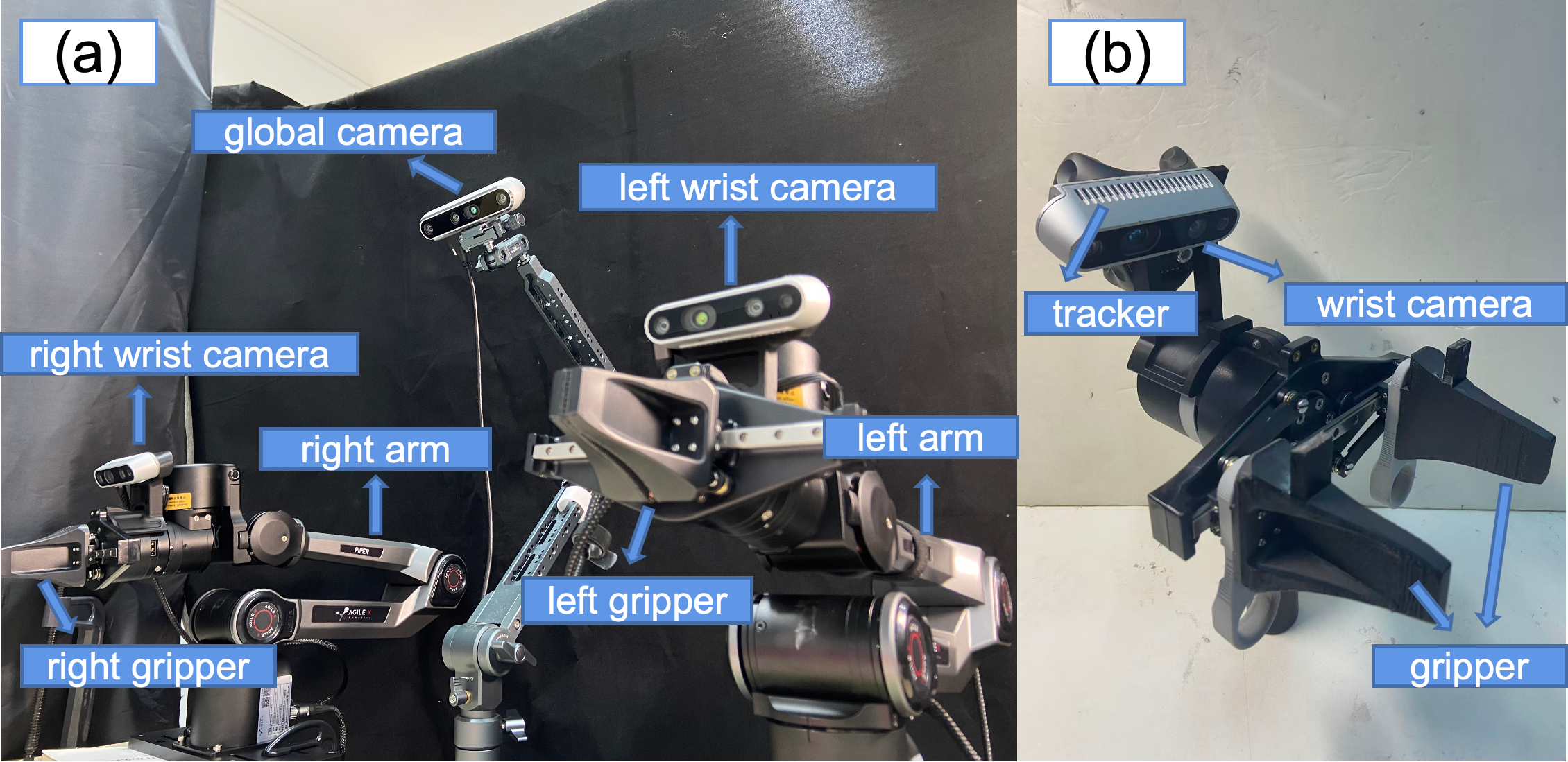}
\vspace{-6pt} 
\caption{(a) Overview of the dual-arm robotic platform, equipped with a global camera, wrist cameras on both arms, and left and right grippers. (b) Close-up view of a UMI gripper, including the wrist camera, tracker, and gripper.}
\label{fig:hardware-setup}
\end{figure}

\subsection{Hardware Setup and Data Acquisition}
The GiT dataset is collected through three pipelines: VR-based real-robot teleoperation, UMI style human demonstration collection, and simulation. The real-robot and simulation pipelines share a dual-arm embodiment comprising two PiperX manipulators, while the UMI style pipeline uses handheld grippers matched to the robot gripper geometry.

\paragraph{VR-based real-robot teleoperation}
The dual-arm platform shown in Fig.~\ref{fig:hardware-setup}(a) is operated using a Meta Quest 3 headset and two handheld controllers through the open-source questVR\_ws \cite{agilex_questvr} framework. The left and right controller poses and button states are acquired through oculus\_reader \cite{orbik2021oculusreader} and mapped to the corresponding robot arms. A ROS-based teleoperation system processes these inputs into Cartesian end-effector targets and gripper commands, with the end-effector targets subsequently converted into joint commands for coordinated bimanual execution.

\paragraph{UMI style human demonstration collection}
This pipeline uses a pair of handheld grippers, one of which is shown in Fig.~\ref{fig:hardware-setup}(b). Instead of the GoPro-based visual-inertial SLAM used in the original UMI \cite{chi2024universal}, pose tracking relies on an external SteamVR system with three VIVE Tracker 3.0 units and two base stations. A table-mounted tracker defines a task-local reference frame, while the other two trackers are mounted on the grippers. Calibrated tracker-to-end-effector transformations convert their poses into end-effector trajectories expressed in this reference frame. Gripper opening is estimated from the image-space separation between two AprilTag markers, normalized using reference measurements at the fully closed and fully open states.

\paragraph{Simulation}
Simulated demonstrations are generated in ManiSkill3 \cite{DBLP:journals/corr/abs-2410-00425} using scripted policies with online motion planning. Each task is decomposed into predefined manipulation primitives---approach, grasp, lift, transport, placement, and release---whose execution is guided by object poses and robot states. Trajectories are generated online using mplib \cite{guo_mplib}, with a screw-based planner for Cartesian motions and an RRT planner for collision-free paths. Bimanual interactions, including object handoffs, are explicitly scripted to reproduce task-specific coordination.

\paragraph{Unified observations and data format}
Both physical pipelines use two Intel RealSense D435i cameras for wrist-view observations and a centrally positioned Intel RealSense D455 camera for a global workspace view. Corresponding views are rendered in simulation, yielding a consistent three-view RGB observation structure across all pipelines. All visual streams are provided at 30 frames per second (fps). Global RGB streams are stored at $1280 \times 720$, while wrist RGB streams are stored at $640 \times 480$ in the real-robot and simulation pipelines and at $1920 \times 1080$ in the UMI style pipeline. Depth observations are additionally recorded in the real-robot and simulation pipelines; no force or tactile measurements are collected. Image sequences and pose trajectories are temporally aligned in software and stored using a unified HDF5 schema.

The GiT dataset comprises 4,680 episodes across 18 tasks, with over 3.3 million synchronized time steps and more than 30 hours of data. Pipeline-specific statistics and stored image resolutions are summarized in Table~\ref{tab:dataset-statistics}.

\begin{table}[!hb]
\centering
\caption{Dataset statistics of the three acquisition pipelines.}
\label{tab:dataset-statistics}
\renewcommand{\arraystretch}{1.12}
\setlength{\tabcolsep}{3.5pt}
\footnotesize
\setlength{\aboverulesep}{0pt}
\setlength{\belowrulesep}{0pt}
\setlength{\extrarowheight}{0.65ex}
\begin{tabular}{lcccc}
\toprule
\textbf{Pipeline}
& \textbf{Tasks}
& \textbf{Episodes}
& \textbf{Time Steps}
& \textbf{Duration (h)} \\
\midrule
Real Robot
& 18 & 1,260 & 639.0K+ & 5.91+ \\
Simulation
& 9 & 1,620 & 1,026.3K+ & 9.50+ \\
UMI
& 18 & 1,800 & 1,681.5K+ & 15.57+ \\
\midrule
\rowcolor{gray!20}
\textbf{Total}
& \textbf{18}
& \textbf{4,680}
& \textbf{3.34M+}
& \textbf{30.99+} \\
\bottomrule
\end{tabular}
\end{table}

\begin {figure*}[ht!]
\centering
\includegraphics [width=\textwidth]{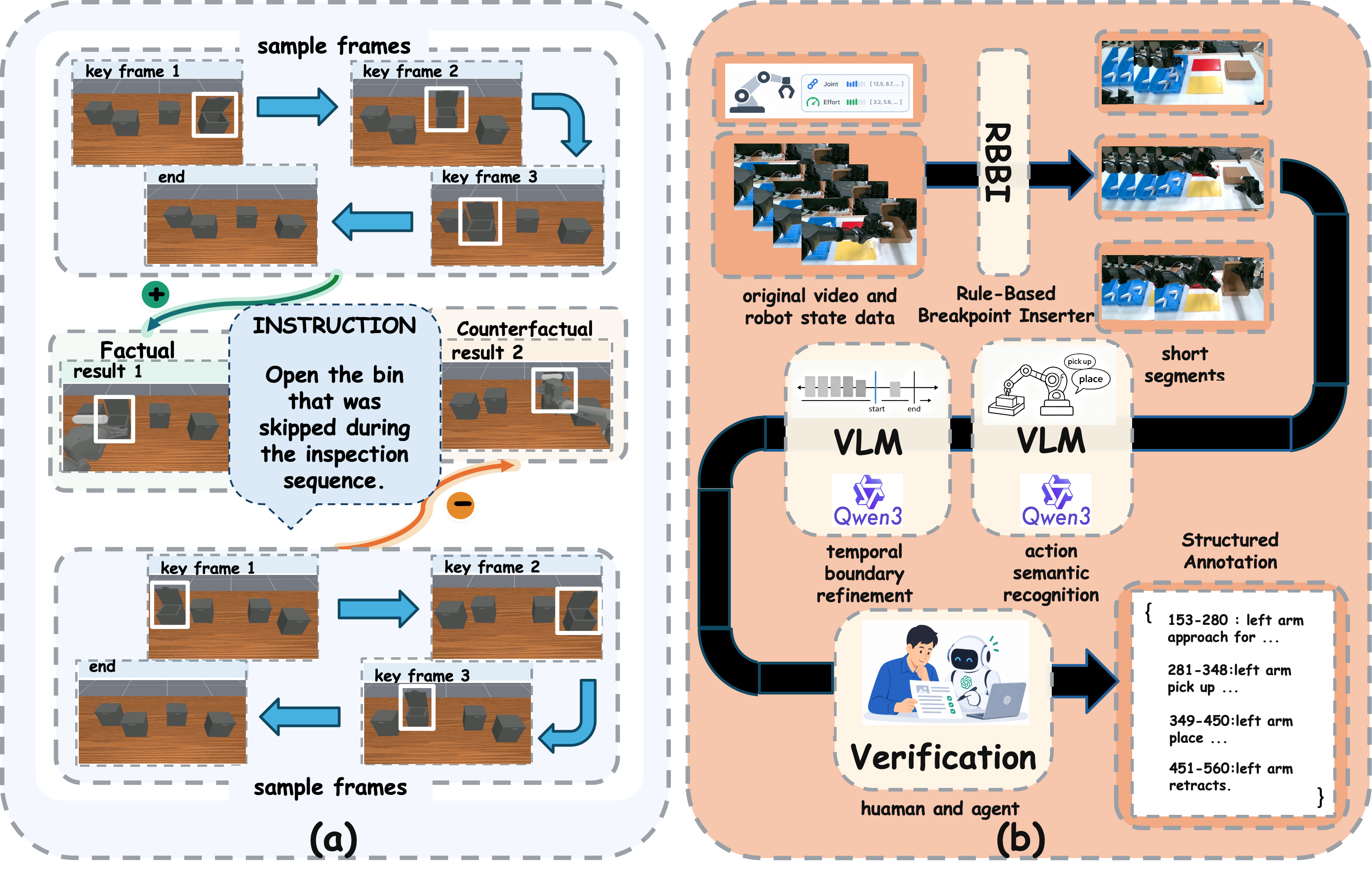}
\caption {(a) \textbf{Counterfactual (CF) pairs.} Each pair shares an identical instruction and an identical final frame but has a completely different demonstration phase, and therefore a completely different correct action. (b) \textbf{Semi-automated annotation pipeline for robotic-arm data.} Videos and robot states processed to identify and refine action segments, which are validated and exported as structured annotations.}
\label {fig:gitbench-overview}
\end {figure*}   

\subsection{Task Design}
The GiT dataset comprises 18 application-oriented manipulation tasks across biolaboratory, household, and industrial settings, with six tasks per setting. The real-robot and UMI style pipelines each cover the complete task suite, spanning diverse workspace layouts, object configurations, and manipulation procedures, as illustrated in Fig.~\ref{fig:git-overview}. Across these tasks, prior interactions determine the referents or requirements of subsequent manipulation instructions.

The GiT benchmark comprises nine of these tasks, with three tasks per setting organized in increasing manipulation complexity. Within each setting, the most complex task requires coordinated bimanual manipulation. This organization provides balanced coverage of the three application settings while assessing pre-episode semantic grounding under different manipulation demands. The simulation pipeline covers the same nine tasks, aligning simulated demonstration collection with the benchmark task suite. Table~\ref{tab:task-overview} summarizes the complete task suite, the required history-dependent capabilities, and the available acquisition pipelines.

We characterize the full task suite using five non-exclusive history-dependent capabilities relevant to \emph{pre-episode semantic grounding}:
(i) \emph{interaction-history recall} (IHR) identifies the manipulation target based on whether or how objects were manipulated in the pre-episode history;
(ii) \emph{object-identity tracking} (OIT) establishes object correspondence between historical and current observations despite motion, occlusion, or rearrangement;
(iii) \emph{temporal-order reasoning} (TOR) uses the relative order of prior interactions to determine the target or required execution order;
(iv) \emph{handoff grounding} (HG) identifies the object involved in a prior handoff as the target of subsequent manipulation; and
(v) \emph{history-guided state restoration} (HSR) requires restoring an object state or scene configuration observed in the pre-episode history.

\subsection{Automated Annotation}
To automatically annotate robot manipulation data, we build a structured action annotation pipeline using raw robot videos and proprioceptive state signals. The overall pipeline is visualized in Fig.~\ref{fig:gitbench-overview}(b). First, the \textit{Rule-Based Breakpoint Inserter (RBBI)} splits long demonstrations into short video segments via heuristic rules on robot state data. We then employ Qwen3-VL-32B\cite{yang2025qwen3}, a vision-language model (VLM), for two-stage reasoning: recognizing action semantics of each segment and refining their temporal boundaries. The pipeline outputs structured annotations with time ranges and action descriptions for manipulation primitives, which can be verified by humans and agents to ensure quality.

\section{Experiments} \label{sec:exp}

\begin {figure*}[ht!]
\centering
\includegraphics [width=\textwidth]{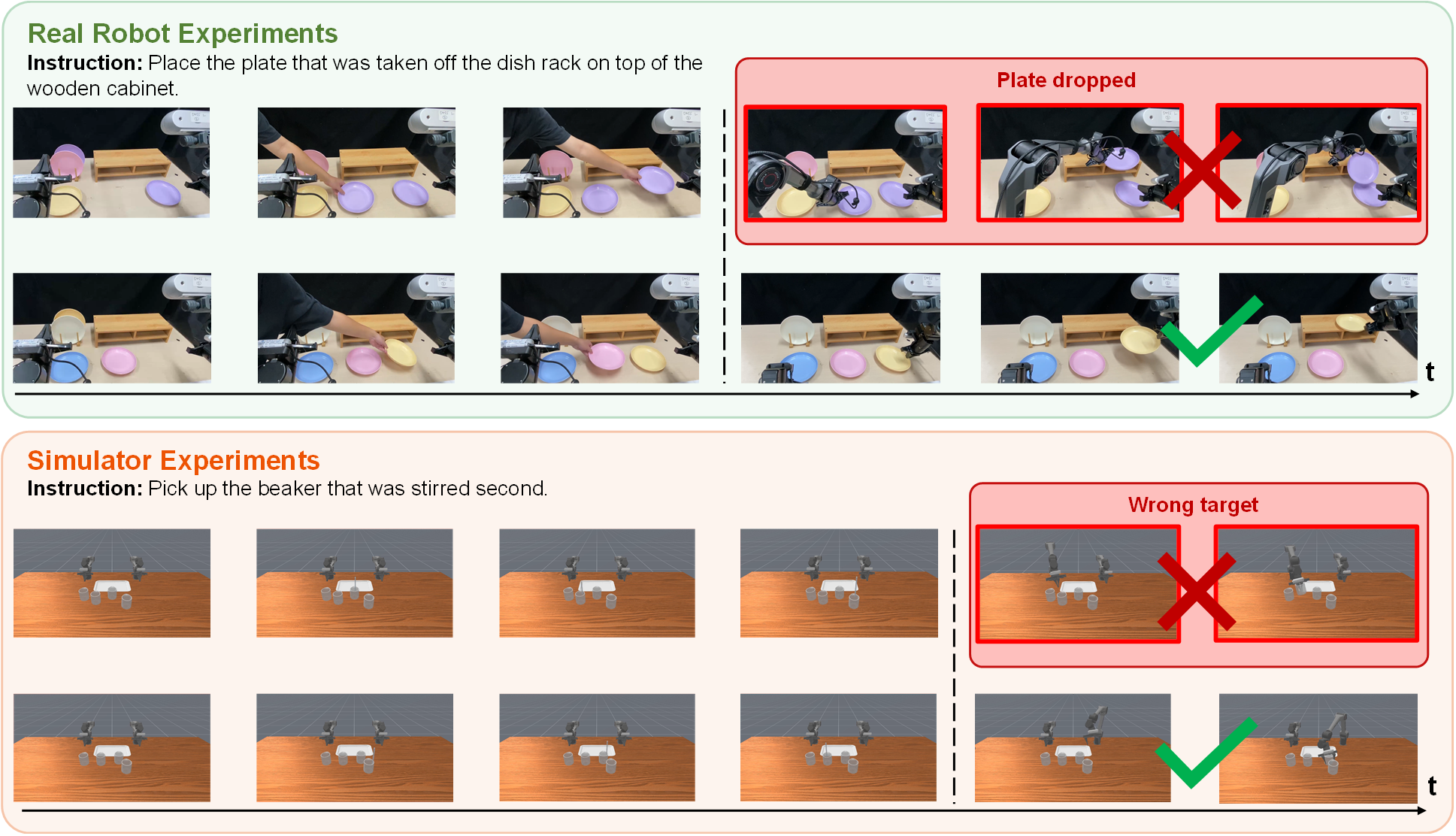}
\caption{
    \textbf{Qualitative examples of successful and failed manipulation in GiT.}
    The top and bottom panels show real-robot and simulated trials, respectively.
    Each panel presents a failed trial in the upper row and a successful trial
    in the lower row, with time progressing from left to right.
    Dashed vertical lines separate pre-episode interaction histories from
    subsequent policy execution.
}
\label {fig:qualitative-results}
\end {figure*}  

\subsection{Experimental Setup}

\paragraph{Data and baselines}
We evaluate $\pi_{0.5}$ \cite{black2025pi05}, RDT \cite{liu2025rdt}, and HAMLET \cite{koo2026hamlet} on the nine simulation tasks of the GiT benchmark, and assess $\pi_{0.5}$ in real-robot experiments. Policies are fine-tuned separately using demonstrations from the corresponding data source: simulated trajectories for simulation and real-robot demonstrations for physical deployment. No UMI demonstrations are used for fine-tuning. All models are initialized from their respective official pretrained weights, with $\pi_{0.5}$ initialized from \texttt{lerobot/pi05\_base}.

\paragraph{Training configuration}
All models undergo full fine-tuning for 30,000 optimization steps using AdamW with an effective batch size of 64. The learning rate increases linearly during a 1,000-step warmup to a peak of $2.5 \times 10^{-5}$, followed by decay to $2.5 \times 10^{-6}$. Each training run uses a single NVIDIA A100 80\,GB GPU with BF16 precision and gradient checkpointing enabled. The action prediction horizon is set to 50 steps.

\paragraph{Evaluation protocol and metric} For each model, we conduct 60 trials per task in simulation and 10 trials per evaluated real-robot task. A trial is successful only if the policy selects the correct object or objects and completes the prescribed manipulation. We report instance-level task success rate (SR):
\begin{equation}
\text{SR}(\pi_\theta) = \frac{1}{M}\sum_{i=1}^{M} S(\pi_\theta, X_i),
\label{eq:sr}
\end{equation}
where $M$ is the number of evaluation trials for a task and $S(\pi_\theta, X_i) \in \{0,1\}$ indicates whether the policy satisfies the success criteria for instance $X_i$. Instances belonging to annotated A–B pairs are scored individually.

For simulation experiments, we further define the process success rate (PSR) to measure intermediate manipulation progress. Instead of only checking final task completion as SR does, PSR uses a staged weighted scoring scheme, which assigns different weights to sequential manipulation sub-steps such as contact, grasp, displacement and target achievement across different task types.
We report instance-level process success rate (PSR):
\begin{equation}
\text{PSR}(\pi_\theta) = \frac{1}{M}\sum_{i=1}^{M} P(\pi_\theta, X_i),
\label{eq:psr}
\end{equation}
where $P(\pi_\theta, X_i) \in \{0,1\}$ marks whether the policy fulfills all required process stages for instance $X_i$.

\begin{table}[htbp]
\centering
\caption{Comparative analysis for action complexity and diversity.}
\label{tab:action_diversity}
\begin{tabular}{llcc}
\toprule
Dataset & Metric & Joint & Rotation \\
\midrule
RoboMME \cite{dai2026robomme} & Complexity & 0.0756 & -- \\
            & Diversity  & 0.2834 & -- \\
GiT (ours) & Complexity & 0.1782 & 0.3468 \\
          & Diversity  & 0.4991 & 0.6825 \\
\bottomrule
\end{tabular}

\vspace{6pt}

\parbox{0.96\textwidth}{
\scriptsize
Joint and rotation is given in radians.
}
\end{table}

\begin{table*}[t]
\centering
\caption{Performance comparison of RDT-1B, $\pi_{0.5}$ and HAMLET on simulation tasks.}
\label{tab:results}
\renewcommand{\arraystretch}{1.08}
\setlength{\tabcolsep}{3.5pt}
\footnotesize
\resizebox{\textwidth}{!}{%
\begin{tabular}{
  @{}l
  c c  c c  c c
  c c  c c  c c
  c c  c c  c c
  c c@{}}
\toprule
\textbf{Method} &
\multicolumn{2}{c}{\textbf{L2}} & \multicolumn{2}{c}{\textbf{L4}} & \multicolumn{2}{c}{\textbf{L5}} &
\multicolumn{2}{c}{\textbf{H1}} & \multicolumn{2}{c}{\textbf{H3}} & \multicolumn{2}{c}{\textbf{H5}} &
\multicolumn{2}{c}{\textbf{I2}} & \multicolumn{2}{c}{\textbf{I3}} & \multicolumn{2}{c}{\textbf{I5}} &
\multicolumn{2}{c}{\textbf{AV}} \\
\cmidrule(lr){2-3} \cmidrule(lr){4-5} \cmidrule(lr){6-7}
\cmidrule(lr){8-9} \cmidrule(lr){10-11} \cmidrule(lr){12-13}
\cmidrule(lr){14-15} \cmidrule(lr){16-17} \cmidrule(lr){18-19}
\cmidrule(lr){20-21}
& SR & PSR & SR & PSR & SR & PSR & SR & PSR & SR & PSR & SR & PSR & SR & PSR & SR & PSR & SR & PSR & SR & PSR \\
\midrule
RDT-1B\cite{liu2025rdt}
& 1.67\% & 4.04\% & 0\% & 7.50\% & 0\% & 0\%
& 1.67\% & 6.50\% & 3.33\% & 3.17\% & 0\% & 5.04\%
& 5.00\% & 7.50\% & 1.67\% & 3.84\% & 0\% & 3.54\%
& 1.48\% & 4.57\% \\
$\pi_{0.5}$\cite{black2025pi05}
& \textbf{5.00}\% & 5.00\% & \textbf{3.33}\% & \textbf{7.50}\% & 0\% & 4.00\%
& 3.33\% & \textbf{8.50}\% & \textbf{1.67}\% & \textbf{3.83}\% & 0\% & \textbf{3.38}\%
& 3.33\% & 8.50\% & \textbf{3.33}\% & 4.27\% & 0\% & 4.00\%
& \textbf{2.22}\% & \textbf{5.44}\% \\
HAMLET\cite{koo2026hamlet}
& 3.33\% & \textbf{5.44}\% & 1.67\% & 4.32\% & \textbf{1.67}\% & \textbf{4.47}\%
& \textbf{5.00}\% & 3.26\% & 0\% & 3.17\% & 0\% & 1.17\%
& \textbf{5.00}\% & \textbf{8.83}\% & 3.33\% & \textbf{5.00}\% & 0\% & \textbf{6.50}\%
& \textbf{2.22}\% & 4.68\% \\
\bottomrule
\end{tabular}}

\vspace{6pt}

\parbox{0.96\textwidth}{
\scriptsize\strut
\textit{SR}: success rate; \textit{PSR}: process success rate; \textit{AV}: average performance over all nine simulation tasks.
}
\end{table*}

\subsection{Quantifying the diversity of the dataset}
We evaluate the action complexity and diversity of our proposed GiT dataset for dua arm manipulation. To quantify action complexity, we calculate the joint state variance across timesteps within each action sequence and average these statistics over all demonstrations to obtain the overall dataset complexity, covering both joint angles and end-effector rotation. For action diversity evaluation, we uniformly sample all action sequences to 300 time ticks. We then compute the cross-sequence variance at each sampled timestamp and average the results across the entire sequence length to measure the trajectory diversity of our robot manipulation demonstrations for both joint angles and rotation.

Quantitative results show that our GiT dataset achieves favorable action complexity and diversity, with a joint-space complexity of 0.1782 rad, rotational complexity of 0.3468 rad, joint-space diversity of 0.4991 rad, and rotational diversity of 0.6825 rad. Compared with RoboMME\cite{dai2026robomme}, our dataset improves joint-space complexity by 135.59\% and joint-space diversity by 76.08\%. Note that RoboMME does not provide Euler-angle annotations in its released data, so a direct comparison on the rotational dimension is not available. These metrics demonstrate that our dataset contains rich variations in robot joint angles and rotation trajectories, covering diverse and complex garment manipulation behaviors. The high variability in joint and rotational trajectories validates that our dataset is well-established for supporting comprehensive research on various garment manipulation scenarios and robotic manipulation learning tasks.

\subsection{Real-Robot Experiments}\label{subsec:result}
We further evaluate temporal grounding capability in real-world manipulation using $\pi_{0.5}$, a representative vision-language-action model that has demonstrated strong performance across a broad range of manipulation benchmarks. We select $\pi_{0.5}$ as the real-robot evaluation model due to its general applicability and competitive performance among recent VLA approaches, providing a strong baseline for assessing history-dependent manipulation in physical environments. Since the original $\pi_{0.5}$ policy performs action prediction based on the current observation rather than explicit historical context, the model does not have pre-episode historical inputs. 

We evaluate $\pi_{0.5}$ on six randomly selected tasks from the three application scenarios. The model achieves varying success rates, with SRs of 10\%, 0\%, 0\%, and 10\% on tasks L1, L4, I2, and I6, respectively, while obtaining relatively higher SRs of 30\% and 40\% on H2 and H3. Figure~\ref{fig:qualitative-results} presents two representative executions of task H2. In both trials, the robot correctly identifies the target object associated with the instruction; however, one execution fails during placement due to object dropping. Despite these successful instances, $\pi_{0.5}$ frequently fails to associate historical interactions with the correct manipulation target across the evaluated tasks. The limited success observed in several tasks suggests that the model can occasionally complete history-dependent instructions, but does not yet exhibit reliable temporal grounding capability. These results highlight the remaining challenges of enabling VLA models to effectively utilize pre-episode interaction history for real-world manipulation.

\subsection{Simulation Experiments}\label{subsec:result}
Table~\ref{tab:results} reports the success rates of RDT-1B,
$\pi_{0.5}$, and HAMLET on the nine simulation tasks of the GiT
benchmark. Under the simulation-only fine-tuning setting,
$\pi_{0.5}$ and HAMLET achieve the same average SR of 2.22\%,
compared with 1.48\% for RDT-1B. Performance remains low across
all tasks, with no model exceeding 5.00\% SR, corresponding to
at most three successful trials out of 60. In particular, none
of the models succeeds on H5 or I5, while HAMLET records only
one successful trial on L5. These results indicate substantial
difficulty in completing the evaluated history-dependent
manipulation tasks. The relative performance of the models varies across tasks. Thus, no evaluated model consistently outperforms the others, and HAMLET does not show a uniform advantage despite incorporating memory mechanisms.

Figure~\ref{fig:qualitative-results} illustrates one source of failure through the instruction ``Pick up the beaker that was stirred second.'' The failed simulated rollout selects an incorrect beaker, whereas the successful rollout identifies and manipulates the intended target. The performance of $\pi_{0.5}$ in simulation was even lower than that observed in the real-world evaluation, which may be partly attributed to the randomness introduced by the limited number of real-world evaluation samples. Although HAMLET incorporates memory mechanisms, it did not demonstrate consistently better performance in our evaluation and was even inferior to $\pi_{0.5}$ in several tasks, potentially because its memory module is not explicitly designed to retrieve and utilize task-relevant information from historical human demonstrations. Overall, these results suggest that existing VLA architectures still have limited capability in achieving semantic grounding based on pre-episode historical information.

\section{Conclusion and Future Work}
We introduced \textbf{Grounded in Time (GiT)}, a multi-source dataset and benchmark for temporal grounding in robotic manipulation, with a focus on \textit{pre-episode semantic grounding} under perceptual ambiguity. The GiT dataset links prior interaction videos to subsequent instructions and execution trajectories across biolaboratory, household, and industrial settings. Evaluations of selected VLA models in simulation, complemented by small-scale real-robot experiments, show that no tested model consistently outperforms the others across the evaluated tasks. All evaluated models exhibit limitations in history-dependent manipulation, highlighting remaining challenges in grounding current instructions in prior interactions.

GiT currently focuses on fixed-workspace bimanual manipulation. Future work will extend the dataset and benchmark to mobile manipulation and longer-horizon workflows requiring multiple memory capabilities, while exploring how different memory representations can be integrated within a unified framework. We hope GiT will advance temporal grounding and, more broadly, research on memory in VLA models.

\bibliographystyle{IEEEtran}
\bibliography{root}

\end{document}